\documentclass[twoside,11pt]{article}

\usepackage[accepted]{melba}  
\usepackage{booktabs}  
\usepackage{wrapfig}
\usepackage{caption}
\usepackage{comment}
\usepackage[T1]{fontenc}
\usepackage{xcolor}
\usepackage{graphicx}
\usepackage{amsmath}
\usepackage{booktabs}
\usepackage{amssymb}
\usepackage{comment}
\usepackage{wrapfig}
\usepackage{todonotes}
\usepackage{subcaption}
\usepackage{multirow}
\usepackage{bm}
\usepackage{hyperref}
\usepackage{amsmath,amsfonts}

\def \Rm{{\mathbb{R}}}

\def \xbf{{\mathbf x}}

\def \ybf{{\mathbf y}}

\def \0bf{{\mathbf 0}}

\melbaheading{2022:005}{https://www.melba-journal.org/papers/2022:005.html}{2022}{1-24}{09/2021}{02/2022}{Raghavendra Selvan and Erik B Dam and Søren Alexander Flensborg and Jens Petersen}{Information Processing in Medical Imaging (IPMI) 2021}{Aasa Feragen, Stefan Sommer, Julia Schnabel, Mads Nielsen}
\ShortHeadings{Patch-based image segmentation with MPS Tensor Networks}{Selvan, Dam,  Flensborg and Petersen}
\firstpageno{1}

\title{Patch-based Medical Image Segmentation using Matrix Product State Tensor Networks}
\author{\name{Raghavendra Selvan} $^{1,2}$ \email{raghav@di.ku.dk}
\AND
\name{Erik B Dam} $^{1}$ \email{erikdam@di.ku.dk}
\AND
\name{{Søren Alexander Flensborg}} $^{1}$ \email{alexanderflensborg@hotmail.dk}
\AND 
\name{Jens Petersen}$^{1,3}$ \email{phup@di.ku.dk}\\ \\
\addr $^{1}$ Department of Computer Science, University of Copenhagen, Denmark\\
\addr $^{2}$ Department of Neuroscience, University of Copenhagen, Denmark \\
\addr $^{3}$ Department of Oncology, Rigshospitalet, Denmark} 

\begin{document}
\maketitle

\begin{abstract}
%

Tensor networks are efficient factorisations of high-dimensional tensors into a network of lower-order tensors. They have been most commonly used to model entanglement in quantum many-body systems and more recently are witnessing increased applications in supervised machine learning. In this work, we formulate image segmentation in a supervised setting with tensor networks. The key idea is to first lift the pixels in image patches to exponentially high-dimensional feature spaces and using a linear decision hyper-plane to classify the input pixels into foreground and background classes. The high-dimensional linear model itself is approximated using the matrix product state (MPS) tensor network. The MPS is weight-shared between the non-overlapping image patches resulting in our {\em strided tensor network} model. The performance of the proposed model is evaluated on three 2D- and one 3D- biomedical imaging datasets. The performance of the proposed tensor network segmentation model is compared with relevant baseline methods. In the 2D experiments, the tensor network model yields competitive performance compared to the baseline methods while being more resource efficient.\footnote{Source code: \url{https://github.com/raghavian/strided-tenet}}\footnote{This is an extension of the preliminary conference work in~\citet{selvan2021segmenting}.}

\end{abstract}

\section{Introduction}

The past decade has witnessed remarkable progress in several key computer vision tasks with the enthusiastic adoption of deep learning methods~\citep{lecun2015deep,schmidhuber2015deep}. Several auxiliary advancements such as powerful hardware for parallel processing (in the form of graphics processing units/GPUs), access to massive amounts of data, better optimizers~\citep{ruder2016overview}, and many practical tricks such as data augmentation, dropout~\citep{hinton2012improving}, batch normalization~\citep{ioffe2015batch} etc. have contributed to the accelerated improvement in the performance of convolutional neural networks (CNNs)-based computer vision models~\citep{lecun1989backpropagation}. 
 Biomedical image segmentation has also been immensely influenced by these developments, in particular with the classic  U-Net~\citep{ronneberger2015u} to the more recent nnU-Net~\citep{isensee2021nnu}.

The dependency of these CNN-based models on high-quality labeled data and specialized hardware could make them less desirable in certain situations. For instance, when dealing with biomedical images where labeled data are expensive and hard to obtain; or in developing countries where access to expensive hardware might be scarce~\citep{ahmed2020democratization}. The carbon footprint associated with training large deep learning models is also of growing concern~\citep{selvan2020carbonfootprint}. Investigating fundamental ideas that can alleviate some or all of these concerns could be valuable going forward. It is in this spirit that we here explore the possibility of using tensor networks for image segmentation.

Tensor networks allow manipulations of higher order tensors in a computationally efficient manner by factorising higher order tensors into networks of lower-order tensors. This property of theirs has been used to study entanglement in quantum many-body systems~\citep{orus2014practical}. From a machine learning point of view, they have similarities with kernel based methods; in that, tensor networks can be used to approximate linear decisions in exponentially high-dimensional spaces~\citep{novikov2018exponential}. This feature of theirs has been used in wide ranging applications such as:  dimensionality reduction~\citep{cichocki2016tensor}, feature extraction~\citep{bengua2015matrix}, to compress neural networks~\citep{novikov2015tensorizing, dai2020video} and discrete probabilistic modelling~\citep{miller2021probabilistic,bonnevie2021mps}. {Other recent works such as in~\citet{dai2020video} perform video scene segmentation using CNNs and employ tensor networks to compress the CNN operations.}
Machine learning using tensor networks is a recent topic that is gaining traction in both -- physics and ML -- communities with works such as~\citet{stoudenmire2016supervised,efthymiou2019tensornetwork}. Tensor network based machine learning offers a novel class of models that have several interesting properties: they are linear, end-to-end learnable and can be resource efficient in many situations~\citep{raghav2020tensor}. They have been successfully used in several imaging based downstream tasks such as generative modelling of small images~\citep{han2018unsupervised} and medical image classification~\citep{selvan2020locally}. 

{In this work, we present the {\em strided tensor network} which is a tensor network based medical image segmentation method. The segmentation is performed on image patches by approximating {\em linear} pixel classification rules using tensor networks. This approach has similarities to other classical pixel classification methods used for segmentation such as ~\citet{soares2006retinal,vermeer2011automated} that operate in a hand-crafted feature space. The key difference of the proposed method is that it does not require designed features that encode domain specific knowledge. Further, the proposed model can be trained in an end-to-end manner in a supervised learning set-up by backpropagating a relevant loss function. This work is based on the preliminary work in~\citet{selvan2021segmenting}, which was the first formulation of tensor networks for image segmentation\footnote{\citet{selvan2021segmenting} was presented at the 27th international conference on Information Processing in Medical Imaging (IPMI), June 28th – June 30th, 2021}.
The key contributions, different from \citet{selvan2021segmenting}, in this work are:
\vspace{-0.5em}
\begin{enumerate}\itemsep-0.5em 
    \item Investigation of tensor network based 3D segmentation method
    \item Novel hybrid segmentation method that combines CNNs and tensor networks
    \item Discussions on different local feature maps
    \item Comprehensive evaluation on three diverse 2D datasets
    \item Experiments on a 3D dataset
    \item Additional baseline methods for comparisons
\end{enumerate}}


%
\begin{figure}[t]
    \centering
    \includegraphics[width=0.9\textwidth]{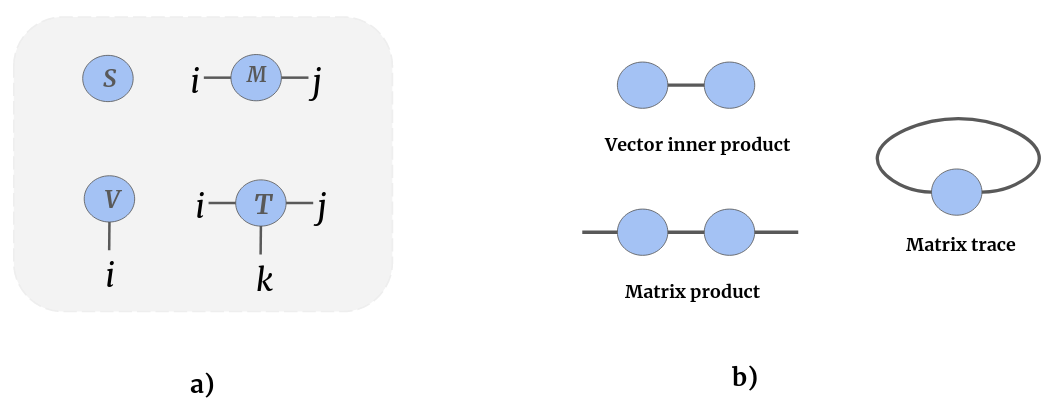}
    \caption{ {\bf a)} Graphical tensor notation depicting an order-0 tensor (scalar $S$), order-1 tensor (vector $V^i$), order-2 tensor (matrix $M^{ij}$) and an order-3 tensor $T^{ijk}$. Tensor indices -- seen as the dangling edges -- are written as superscripts by convention. {\bf b)} Tensor notations can capture operations involving higher order tensors succinctly. For instance, vector inner product is depicted on top where there is a single common edge over which the summation is carried out. Similarly, matrix product of two matrices is shown where the common edge is subsumed in the operation. As trace of a matrix yields a scalar this is depicted as the self-connected edge in the final tensor diagram.}
    \label{fig:tensor}
\end{figure}

\section{Background}
Working with tensors of orders higher than three can be cumbersome. Tensor notations are graphical representations of high-dimensional tensors and operations on them as introduced in~\citep{penrose1971applications}. A grammar of tensor notations has evolved through the years enabling representation of complex tensor algebra. This not only makes working with high-dimensional tensors easier but can also provides insight into efficiently manipulating them. Figure~\ref{fig:tensor} shows the basics of tensor notations and one important operation -- tensor contraction (in sub-figure b). We build upon the ideas of tensor contractions to understand tensor networks such as the MPS, which is used extensively in this work. We point to resources such ~\citep{bridgeman2017hand} for more detailed introduction to tensor notations.
\footnote{\url{https://tensornetwork.org/} also has some well-curated introductory material.}

Tensor networks for image classification tasks have been studied extensively in recent years. The earliest of the image classification methods that used tensor networks in a supervised learning set-up is the seminal work in~\citet{stoudenmire2016supervised}. This work was followed up by several other formulations of supervised image classification with tensor networks~\citep{klus2019tensor,efthymiou2019tensornetwork}. The key difference between these methods is in the procedures used to optimise parameters of the tensor network, and the 1D input representations used. The methods in~\citet{stoudenmire2016supervised,klus2019tensor} use the density matrix renormalisation group (DMRG)~\citep{schollwock2005density,mcculloch2007density} whereas in ~\citet{efthymiou2019tensornetwork} optimisation is performed using automatic differentiation. The above mentioned methods use the matrix product state (MPS) tensor network which is defined for 1D inputs, and they use a simple flattening operation to convert small, 2D images into 1D vector input to MPS. { {Additional modifications to tensor network based methods were required for dealing with more complex and higher resolution image data, such as the ones encountered in medical image analysis. In~\citet{raghav2020tensor,selvan2020locally}, a multi-layered tensor network approach was used where instead of a single flattening operation smaller image neighbourhoods were {\em squeezed} to retain spatial structure in the image data.}}
\begin{figure}[t]
    \centering
    \includegraphics[width=0.99\textwidth]{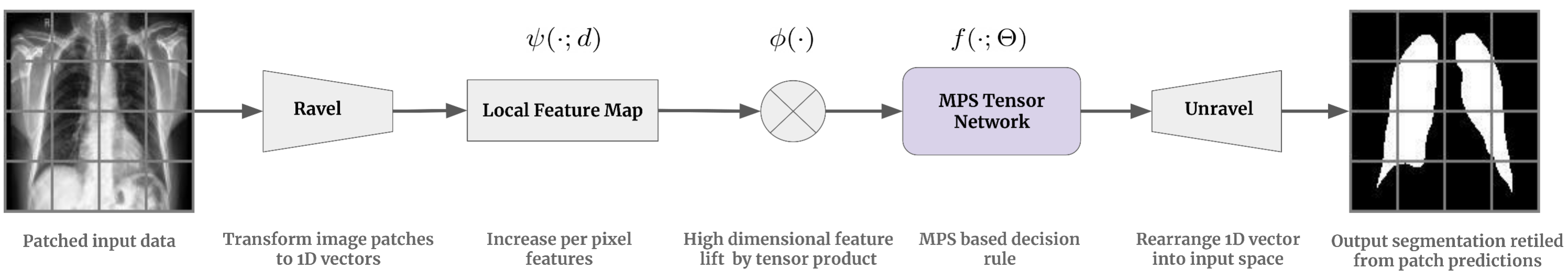}
    
    \caption{{\bf a)} High level overview of the proposed strided tensor network model. Matrix product state (MPS) operations are performed on non-overlapping regions of the input image, $X$, of size H$\times$W resulting in the predicted segmentation. The flattening operation is depicted as "ravel" which flattens 2D $K\times K$image patches into 1D vectors of size $K^2$. Local feature maps $\psi(\cdot)$ are applied to individual pixels {(elements of 1D vector)} to increase the local features. The global feature map, $\phi(\cdot)$, is obtained by performing the tensor outer product of local feature maps. A matrix product state (MPS) tensor network with learnable parameters, $\Theta$, is used to learn a linear segmentation model. Finally, the 1D segmentation maps are transformed back into the image space (unravel operation) to obtain the predicted segmentation.} 
    
    \label{fig:model}
    
\end{figure}


\section{Methods}

\subsection{Overview}
We propose a tensor network based model to perform image segmentation. This is performed by learning a linear segmentation model in an exponentially high-dimensional feature space. Specifically, we derive a hyper-plane such that is able to classify pixels in a high-dimensional feature space into foreground and background classes across all image patches in the dataset. A high level overview of the proposed model that describes these steps is illustrated in Figure~\ref{fig:model}.

Input images are first converted into non-overlapping image patches. {These image patches are flattened (ravel operation in Figure~\ref{fig:model}) into 1D vectors and simple feature transformations such as the sinusoidal transformations in~\citet{stoudenmire2016supervised} are applied to each pixel {(elements of 1D vector)} to obtain the {\em local feature maps}. These local feature maps are similar to the feature extraction performed in other model-based methods that use multi-scale Hessian features in vesselness-type methods~\cite{frangi1998multiscale},
except that in tensor networks these local feature maps are used in lifting the input data to a higher dimensional space.} The lift itself is achieved by performing tensor products of the local feature maps resulting in the {\em global feature map}. Weights of  a linear model that operates on the global feature map are approximated using the matrix product state (MPS) tensor network\footnote{Matrix product state tensor networks are also known as Tensor Trains in literature.}~\citep{perez2006matrix,oseledets2011tensor}. These learnable MPS tensor networks are weight-shared across the different image patches (implemented as a batched operation) resulting in our strided tensor network model for performing image segmentation. The predicted segmentations are 1D vectors which are then reshaped (unravel block in Figure~\ref{fig:model}) back into the image space to correspond to the input image patch. These predicted segmentations are compared with the segmentation mask from training data and a suitable loss is backpropagated to optimise the weights of our model. 

We next describe the specifics of approximating the segmentation rule with MPS and a discussion on different local feature maps in the remainder of this section.

\subsection{Image segmentation using linear models}

We start out by describing the image segmentation model that operates on the full input image which will also help highlight some of the challenges of using MPS tensor networks for image segmentation.

Consider a 2 dimensional image, $X \in \Rm^{H\times W \times C}$ with $N=H\times W$ pixels and $C$ channels. The task of obtaining an $M$--class segmentation, $Y \in \{0,1\}^{H\times W \times M}$ is to learn the decision rule of the form $f (\cdot\;;\;\Theta): X \mapsto Y$, which is parameterised by $\Theta$. These decision rules, $f(\cdot\;;\;\Theta)$, could be linear models such as support vector machines~\citep{vapnik1995support} or non-linear transfer functions such as neural networks. In this work, we explore $f(\cdot\;;\;\Theta)$ that are linear and learned using tensor networks. For simplicity, we assume a two class segmentation problem\footnote{The case when M=2 can also be modeled using a single prediction channel (M=1) and applying a p=0.5 threshold as a decision rule per pixel, as is common when using sigmoid- and softmax- activation functions in deep learning.}, implying $M=2$. However, extending this work to multi-class segmentation of inputs is straightforward.

The linear decision rule is not applied to the raw input data but to the input data lifted to an exponentially high-dimensional feature space. This is based on the insight that non-linearly separable data in lower dimensions could become linearly separable when lifted to a sufficiently high-dimensional space~\citep{cortes1995support}. The lift in this work is accomplished in two steps.
\\
{\bf Step 1:} The input image is flattened into a 1-dimensional vector $\xbf \in \Rm^{N\times C}$. Simple non-linear feature transformations, such as the sinusoidal transformations in Eq.~\eqref{eq:local},  are applied to {the elements of the 1D vector (originally image pixels)} to obtain local feature maps. The local feature map for a pixel $x_j$ is given as $\psi^{i_j}(x_j): \Rm^C \mapsto \Rm^{C\cdot d}$ indicating that the local feature maps are applied to each channel of the input pixel. Note that the tensor index $i_j=C\cdot d$ and is the tensor index dimension. Different choices for the local feature maps are presented in Section~\ref{sec:local}.
\\
{\bf Step 2:} A global feature map is obtained by performing the tensor product\footnote{Tensor product is the generalisation of matrix outer product to higher order tensors.} of the local feature maps. This operation takes $N$  order-1 tensors and outputs a single order-$N$ tensor, $\Phi^{i_1\dots i_N}(\xbf) \in \Rm^{(C\cdot d)^N}$ and is given by
\begin{equation}
    \Phi^{i_1\dots i_N}(\xbf) = \psi^{i_1}(x_1) \otimes \psi^{i_2}(x_2) \otimes \dots \otimes \psi^{i_N}(x_N).
    \label{eq:joint}
\end{equation}
Note that after this operation each image can be treated as a vector in the $(C\cdot d)^N$ dimensional Hilbert space~\citep{orus2014practical,stoudenmire2016supervised}.

Given the $(C\cdot d)^N$ global feature map in Eq.~\eqref{eq:joint}, a linear decision function $f(\cdot;\Theta)$ can be estimated by simply taking the tensor dot product of the global feature map with an order-(N+1) weight tensor, $\Theta^{m}_{i_1\dots i_N}$:
\begin{equation}
    f^m(\xbf\;;\;\Theta) =  \Theta^{m}_{i_1\dots i_N}  \cdot  \Phi_{i_1\dots i_N}(\xbf).
    \label{eq:linModel}
\end{equation}
This dot product between an order N and order-(N+1) tensor yields an order-1 tensor as the output (seen here as the tensor index $m$). The resulting order-1 tensor from Eq.~\eqref{eq:linModel} has $N$ entries corresponding to the pixel level segmentations, which can be unraveled back into the image space. Eq.~\eqref{eq:linModel} is depicted in tensor notation in Figure~\ref{fig:mps}-a.

\subsection{Choice of local feature maps}
\label{sec:local}
The first step after flattening the input image is to increase the number of local features per pixel. There are several local maps in literature to choose from: 
\begin{enumerate}
    \item In quantum physics applications, simple sinusoidal transformations that are related to wavefunction superpositions are used as the local feature maps. A general sinusoidal local feature map from~\citep{stoudenmire2016supervised}, which increases local features of a pixel from $1$ to $d$ is given as: 
\begin{equation}
\psi^{i_j}(x_j) =  \sqrt{{d-1 \choose i_j-1}} \left(\cos(\frac{\pi}{2}x_j)\right)^{(d-i_j)}\left(\sin(\frac{\pi}{2}x_j)\right)^{(i_j-1)}
{ \forall \; i_j=1\dots d}. \label{eq:local}
\end{equation}
The intensity values of individual pixels, $x_j$, are assumed to be normalised to be in $[0,1]$. 
\item The feature map introduced in~\citet{efthymiou2019tensornetwork} simply uses intensity-based local feature maps  with $d=2$:
\begin{equation}
\psi^{i_j}(x_j) = [x_j,(1-x_j)].    
\end{equation}
\item Scalable Fourier features used in transformer-type models to encode positional information can also be used to obtain expressive local feature maps~\citep{mildenhall2020nerf,jaegle2021perceiver}:
\begin{equation}
\psi^{i_j}(x_j) =  \left[\sin(2^{i_j}{\pi}x_j), \cos(2^{i_j}{\pi}x_j)\right]
{ \forall \; i_j=1\dots d/2}. \label{eq:local_1}
\end{equation}
\item Finally, one could use a CNN-based feature extractor that operates on the 2D input image to learn a set of local features. Using these learnable local feature maps results in the {\em hybrid strided tensor network} model that combines CNNs for local feature extraction and MPS for segmentation. In this case the number of filters of the final CNN layer is the local feature map dimension, $d$.
\end{enumerate}


\subsection{Strided tensor networks}

The dot product in Eq.~\eqref{eq:linModel} looks conceptually easy, however, working with it unveils two crucial challenges:
\begin{enumerate}\itemsep-0.5em 
    \item Intractability of the dot product
    \item Loss of spatial correlation
\end{enumerate}
We next address each of these challenges along with the proposed solutions.
\subsubsection{Intractability of the Dot Product {addressed by} Matrix Product States}
The number of parameters in the weight tensor $\Theta$ in Eq.~\eqref{eq:linModel} is $(C\cdot d)^N$, which is massive;  even for binary segmentation tasks it can be a really large number. To bring that into perspective, consider the weight tensor required to operate on a single channel, tiny input image of size 16$\times$16 with local feature map $d=2$: the number of parameters in $\Theta$ is $d^N= 2^{256}\approx 10^{77}$ which is close to the number of atoms in the observable universe!\footnote{\url{https://en.wikipedia.org/wiki/Observable_universe}} (estimated to be about $10^{80}$). 

Recollect that the global feature map $\Phi^{i_1\dots i_N}(\xbf)$ in Eq.~\eqref{eq:joint}  is obtained by flattening the input data. Operating in the $(C\cdot d)^N$ dimensional space implies that all the pixels are {\em entangled} or from a graphical model point of view connected to each other (fully connected neighbourhood per pixel), which is prohibitively expensive and inefficient. From an image analysis point of view it is well understood that for downstream tasks such as segmentation, useful pixel level decisions can be performed based on smaller neighbourhoods. This insight that reducing the pixel interactions to a smaller neighbourhood can yield reasonable decision rules leads us to seek out an approximation strategy that accesses only useful interactions in the global feature space.
%
This approximation is only made inevitable as the exact evaluation of the linear model in Eq.~\eqref{eq:linModel} is simply infeasible. 

%
\begin{figure}[h]
    \centering
    \includegraphics[width=0.6\textwidth]{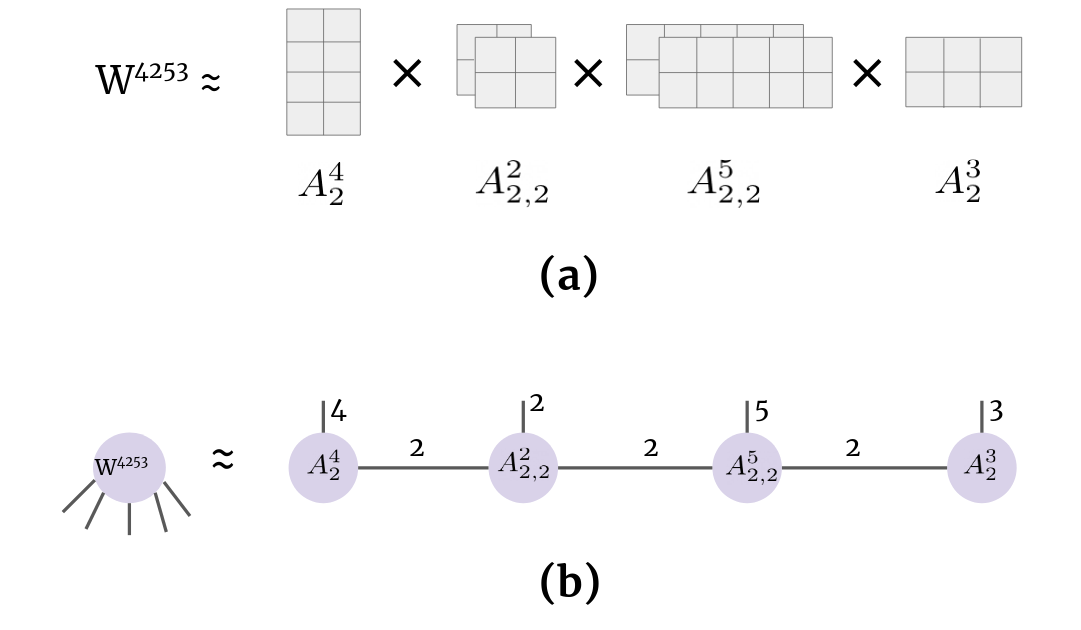}
    \caption{a) An example showing the matrix product state (MPS) factorisation. Here the higher order tensor is $W^{4253}$ of order 4 which is approximated using a network of order-2 and order-3 tensors as shown in the illustration. Performing this factorisation can yield an efficient approximation to $W^{4253}$ comprising the learnable parameters in $A^{i_j}_{\alpha_j \alpha_{j+1}}$. A clear advantage is that the number of parameters in $|W^{4253}| = 120$, whereas the MPS approximation would only comprise $42$ parameters. For higher order tensors the efficiency in reducing the number of parameters can be much higher. b) The MPS factorisation in tensor notation.}
    \label{fig:mps_ex}
\end{figure}


Computing the inner product in Eq.~\eqref{eq:linModel} becomes infeasible with increasing $N$~\citep{stoudenmire2016supervised}. It also turns out that only a small number of degrees of freedom in these exponentially high-dimensional Hilbert spaces are relevant~\citep{poulin2011quantum,orus2014practical}. These relevant degrees of freedom can be efficiently accessed using tensor networks such as the MPS~\citep{perez2006matrix,oseledets2011tensor}. One can think of this as a form of lower dimensional sub-space of the high-dimensional Hilbert space accessed using tensor networks where the task-specific information is present. In image analysis,  accessing this smaller sub-space of the joint feature space could correspond to accessing interactions between pixels that are local either in spatial- or in some feature-space sense that is relevant for the task.

\begin{figure}
    \centering
    \includegraphics[width=0.55\textwidth]{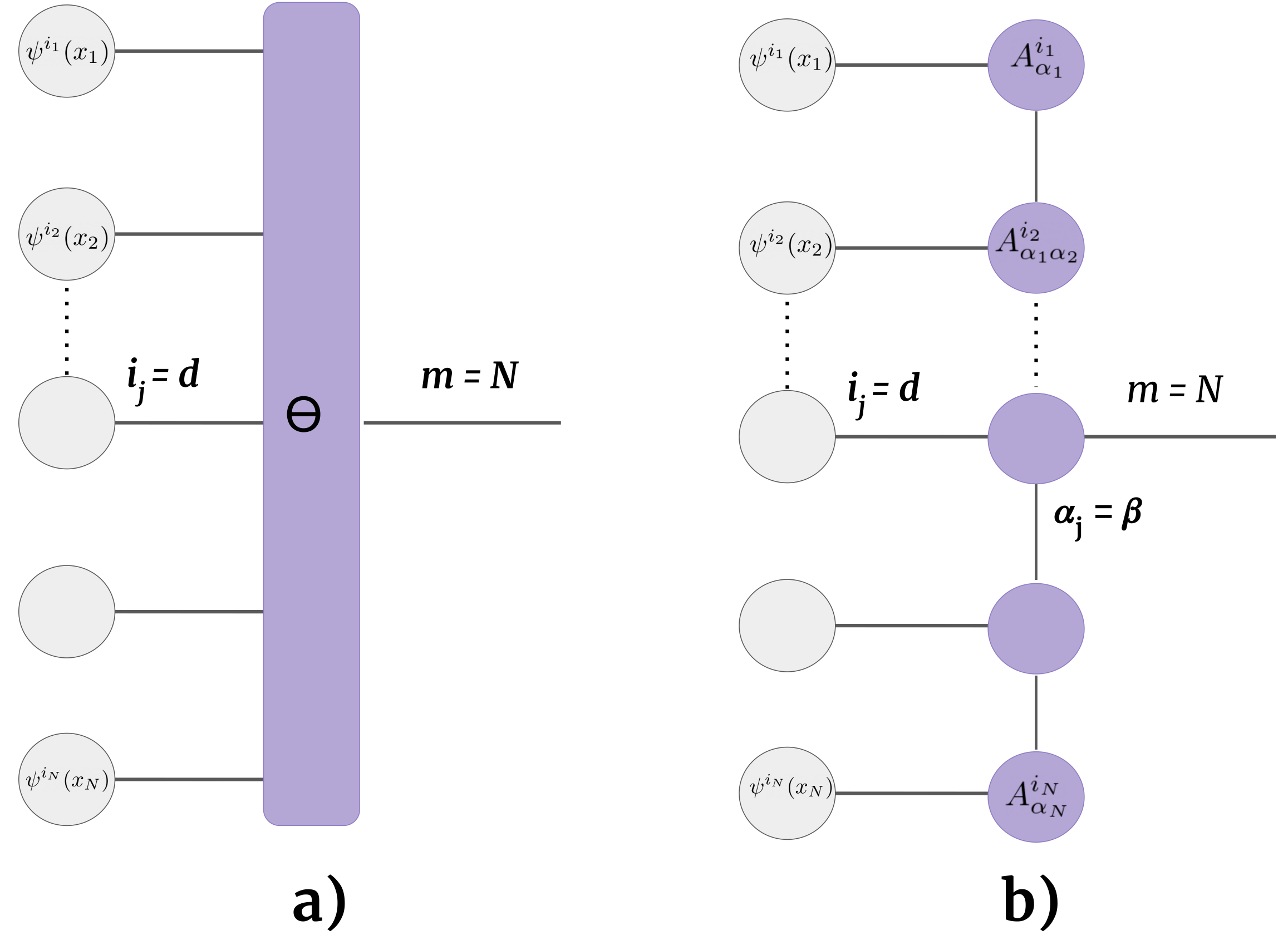}
    \caption{  {\bf a)} Linear decision rule in Eq.~\ref{eq:linModel} depicted in tensor notation. Note that $\Theta$ has N+1 edges as it is an order-(N+1) tensor. The d-dimensional local feature maps are the gray nodes marked $\psi^{i_j}(x_j)$. {\bf b)} Matrix product state (MPS) approximation of $\Theta$ in Eq.~\ref{eq:mps} into a tensor train comprising up to order-3 tensors, $A^{i_j}_{\alpha_j\alpha_{j+1}}$.}
    \label{fig:mps}
\end{figure}

MPS is a tensor factorisation method that can approximate any order-N tensor with a chain of order-3 tensors. This is visualized using tensor notation in Figure~\ref{fig:mps_ex} for an illustrative example. Figure~\ref{fig:mps}-b  shows the tensor notation of the MPS approximation to $\Theta^{m}_{i_1\dots i_N}$ using $A^{i_j}_{\alpha_j \alpha_{j+1}} \forall\; j=1\dots N$ which are of order-3 (except on the borders where they are of order-2). The dimension of subscript indices of  ${\alpha_j}$ which are contracted can be varied to yield better approximations. These variable dimensions of the intermediate tensors in MPS are known as bond dimension $\beta$. MPS approximation of $\Theta^{m}_{i_1\dots i_N}$ depicted in Figure~\ref{fig:mps}-b is given by
\begin{equation}
    \Theta^{m}_{i_1\dots i_N} = \sum_{\alpha_1, \alpha_2,\dots \alpha_N} A^{i_1}_{\alpha_1} A^{i_2}_{\alpha_1 \alpha_2} A^{i_3}_{\alpha_2 \alpha_3} \dots A^{m,i_j}_{\alpha_j \alpha_{j+1}} \dots A^{i_N}_{\alpha_N}.
    \label{eq:mps}
\end{equation}
The components of these intermediate lower-order tensors $A^{i_j}_{\alpha_j \alpha_{j+1}} \forall\; j=1\dots N$ form the tunable parameters of the MPS tensor network. 
This MPS factorisation in Eq.~\eqref{eq:mps} reduces the number of parameters to represent $\Theta$ from $(C\cdot d)^N$ to  $\{N \cdot d \cdot C \cdot \beta^2\}$ with $\beta$ controlling the quality of these approximations\footnote{Tensor indices are dropped for brevity in the remainder of the manuscript.}. Note that when $\beta=(C\cdot d)^{N/2}$ the MPS approximation is exact~\citep{orus2014practical,stoudenmire2016supervised}. 
%
\subsubsection{Loss of Spatial Correlation {addressed by} MPS on Patches} 
The quality of the approximations of the linear model in Eq.~\eqref{eq:linModel} using MPS is controlled by bond dimension $\beta$. As mentioned above the number of parameters grow quadratically with $\beta$ and linearly with the number of pixels. This results in a scenario where the MPS approximation works best for smaller images. This is the case in existing MPS based image analysis methods, where almost always the input image is about $32\times 32$px~\citep{stoudenmire2016supervised,efthymiou2019tensornetwork,han2018unsupervised} and bond dimensions are $<50$. For larger input data and smaller bond dimensions, the necessary spatial structure of the images to perform reasonable approximations of decisions rules might not be retained. And lack of spatial pixel correlation can be detrimental to downstream tasks; more so when dealing with complex structures encountered in medical images. We propose to alleviate this by using a patch-based approach.
The issue of loss in spatial pixel correlation is not alleviated with MPS as it operates on flattened input images. MPS with higher bond dimensions could possibly allow interactions between all pixels but the quadratic increase in number of parameters with the bond dimension $\beta$, working with higher bond dimensions can be prohibitive.

To address this issue, we apply MPS on small non-overlapping image regions. These smaller patches can be flattened with less severe degradation of  spatial correlation. Similar strategy of using MPS on small image regions has been used for image classification using tensor networks in~\citep{raghav2020tensor}. This is also in the same spirit of using convolutional filter kernels in CNNs when the kernel width is set to be equal to the stride length. This formulation of using MPS on regions of size $K\times K$ with a stride equal to $K$ in both dimensions results in the strided tensor network formulation, given as
\begin{align}
f(\xbf;\Theta_K) &= \{ \Theta_K \cdot \Phi( \xbf_{(i,j)})\} \quad \forall\; i = 1,\dots,H/K, \; j = 1,\dots,W/K
    \label{eq:model}    
\end{align}
where $(i,j)$ are used to index the patch from row $i$ and column $j$ of the image grid with patches of size $K\times K$. The weight matrix in Eq.~\eqref{eq:model}, $\Theta_K$ is subscripted with $K$ to indicate that MPS operations are performed on $K\times K$ patches.

In summary, with the proposed strided tensor network formulation, linear segmentation decisions in Eq.~\eqref{eq:linModel} are approximated at the patch level using MPS. The resulting patch level predictions are tiled back to obtain the $H\times W$ segmentation mask. 

\subsection{Optimisation}

The weight tensor, $\Theta_K$, in Eq.~\eqref{eq:model} and in turn the lower-order tensors in Eq.~\eqref{eq:mps} which are the model parameters can be learned in a supervised setting. For a given labelled training set with $T$ data points, $\mathcal{D}:\{(\xbf_1,\ybf_1), \dots (\xbf_T,\ybf_T)\}$, the training loss to be minimised is
\begin{equation}
    \mathcal{L}_{tr} = \frac{1}{T} \sum_{t=1}^T L(f(\xbf_i),\ybf_i),
\end{equation}
where $\ybf_i$ are the binary ground truth masks and $L(\cdot)$ can be a suitable loss function suitable for segmentation tasks. In this work, we use either cross-entropy loss or Dice loss depending on the dataset.

\section{Data \& Experiments}
\label{sec:data_exp}

\subsection{Data}
\label{sec:data}
We report experiments on four medical image segmentation datasets of different image modalities with varying complexities. Three of the datasets are for 2D segmentation and the final one is for 3D segmentation. The datasets are further described below.

\subsubsection{ MO-NuSeg Dataset}
\begin{figure}[t]
    \centering
    \includegraphics[width=0.65\textwidth]{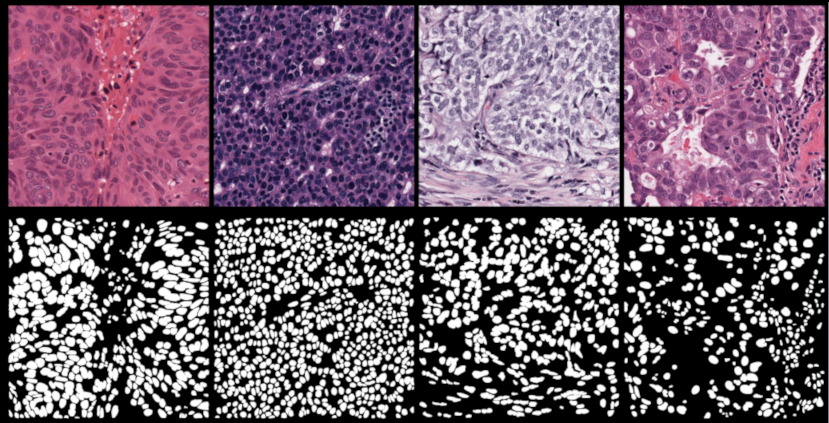}
    
    \caption{Histopathology images from multiple organs (top row) and the corresponding binary masks (bottom row) from the MO-NuSeg dataset~\citep{kumar2017dataset}.}
    \label{fig:monuseg}
\end{figure}

We use the multi-organ nuclei segmentation (MO-NuSeg) challenge dataset\footnote{\url{https://monuseg.grand-challenge.org/}}~\citep{kumar2017dataset} consisting of 44 Hematoxylin and eosin (H\&E) stained tissue images, of size 1000$\times$1000, with about 29,000 manually annotated nuclear boundaries. This dataset is challenging due to the variations in the tissues that are from seven different organs. The dataset consists of 44 images split into 30 training/validation images and 14 testing images. Samples from the MO-NuSeg dataset and their corresponding binary nuclei masks are shown in Figure~\ref{fig:monuseg}.
\label{sec:lungCXR}
\begin{figure}[h]
    \centering
    \includegraphics[width=0.65\textwidth]{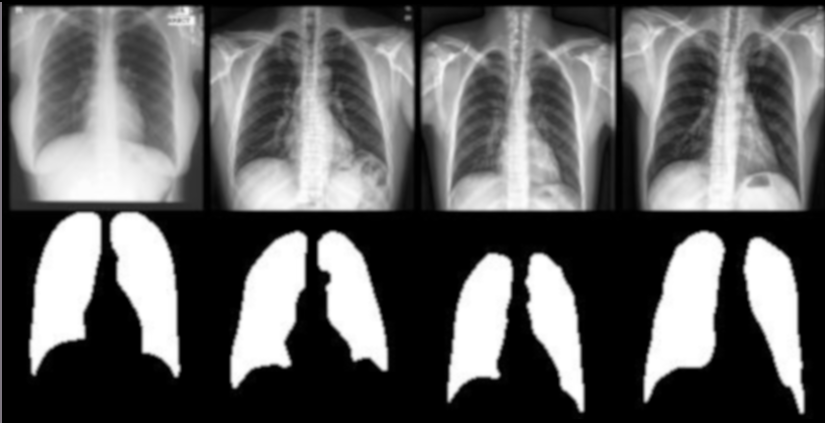}
    
    \caption{ Four chest X-ray images (top row) and the corresponding binary masks (bottom row) from the Lung-CXR dataset. }
    \label{fig:cxr}
\end{figure}
\subsubsection{Lung-CXR Dataset}

The lung chest X-ray dataset is collected from the Shenzhen and Montgomery hospitals with posterio-anterior views for tuberculosis diagnosis~\citep{jaeger2014two}. The CXR images used in this work are scaled down to 128$\times$128 with corresponding binary lung masks for a total of 704 cases which is split into training ($352$), validation ($176$) and test ($176$) sets. Four sample CXRs and the corresponding lung masks are shown in the Figure~\ref{fig:cxr}.

\subsubsection{Retinal vessel segmentation dataset}
\begin{figure}[t]
    \centering
    \includegraphics[width=0.65\textwidth]{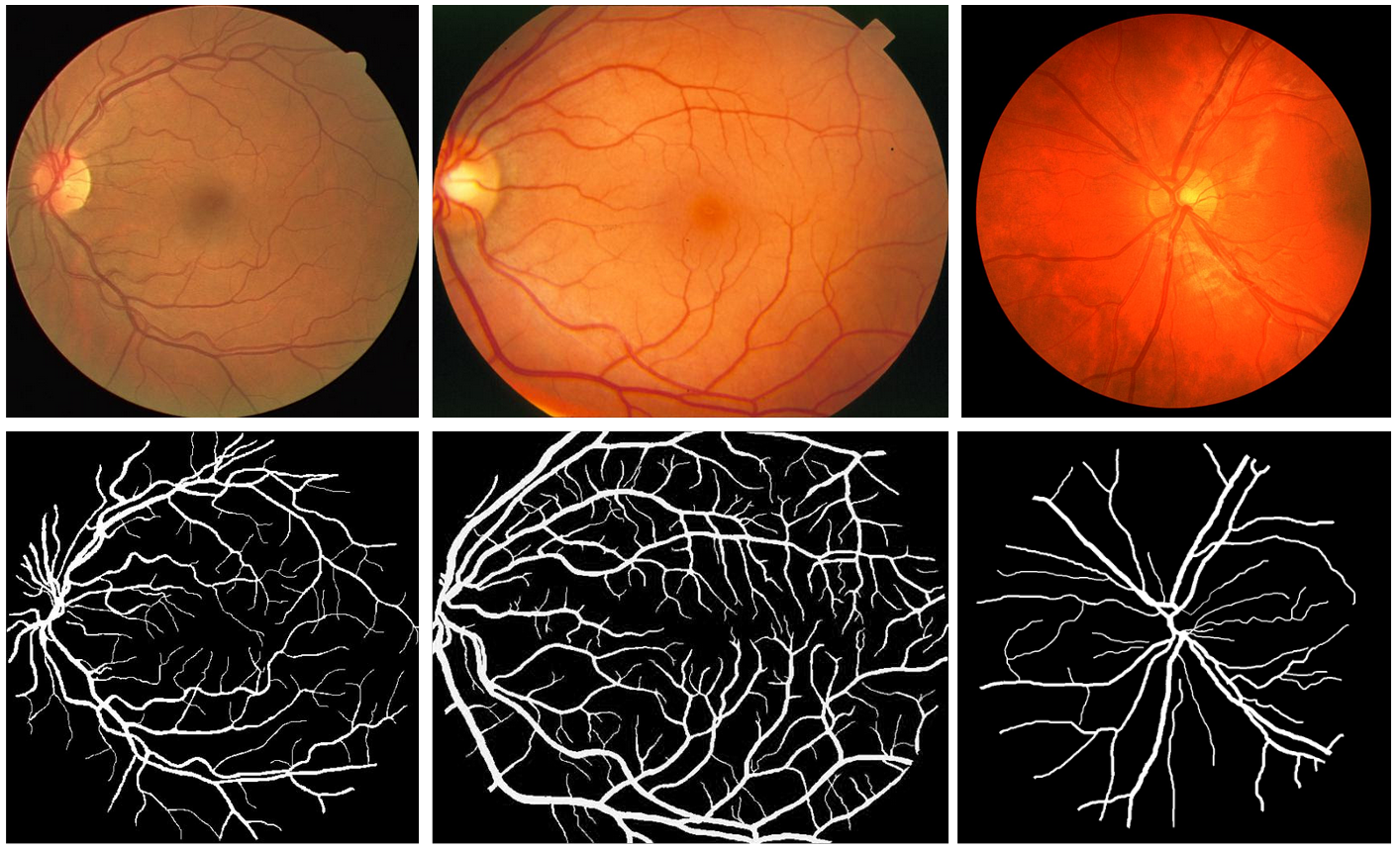}
    
    \caption{Sample RGB images from the retinal vessel segmentation dataset. The variations in the dataset due to the underlying datasets (DRIVE, STARE and CHASE) is evident in these samples. Binary vessel segmentations are visualized in the bottom row.}
    \label{fig:retina}
\end{figure}

The third 2D segmentation task was formulated by pooling three popular retinal vessel segmentation datasets: DRIVE~\citep{staal2004ridge}, STARE~\citep{,hoover2000locating} and CHASE~\citep{owen2011retinal}. In total we obtained 68 RGB images of height 512 px and of width varying between 512px and 620px. A random split of 48+10+10 images was used for training, validation and test purposes, respectively. Sample images along with corresponding vessel segmentation masks from the retinal dataset are shown in Figure~\ref{fig:retina}.

\subsubsection{Brain tumour dataset}
\begin{figure}[h]
    \centering
    \includegraphics[width=0.65\textwidth]{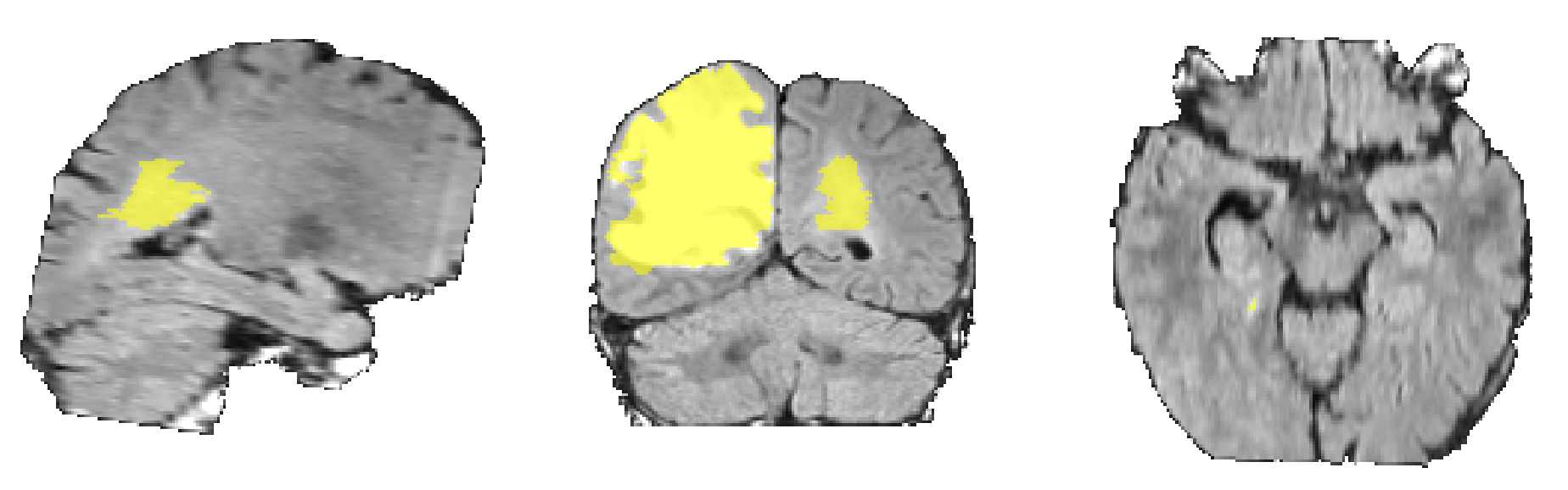}
    
    \caption{Three views of the T1 volume for one of the BraTS dataset volumes are visualized above with an overlay of the tumour regions (yellow).~\citep{brats2014multimodal}}
    \label{fig:brats}
\end{figure}

The multimodal brain tumour segmentation (BraTS) dataset~\citep{brats2014multimodal,brats2017advancing,brats2018identifying} from the 2016 and 2017 challenge edition is used to study the segmentation performance on 3D data. The BraTS dataset consists of 3D volumes of skull stripped brain images in four modalities: T1, post-contrast T1-weighted, T2-weighted and T2 fluid attenuated inversion recovery (FLAIR), acquired with different scanners from multiple institutions. Manual annotations for three tumour sub-types are provided; in this work we create a binary segmentation task by merging the three classes. The total dataset used in this work consists of 484 4D volumes obtained from the Medical Decathlon challenge~\citep{antonelli2021medical}\footnote{\url{http://medicaldecathlon.com/}} which are split into three partitions for training (242), validation (121) and test (121) purposes. All volumes were of size 224x224x160 with isotropic 1mm$^3$ voxels. Three views of one of the data points is visualized in Figure~\ref{fig:brats} where the tumour regions are overlaid with the T1 volume.

\subsection{Experiments}
\label{sec:exp}

\subsubsection{Experimental set-up} 

The experiments were designed to study the segmentation performance of the proposed strided tensor network model on the three 2D- and one 3D- datasets. Different, relevant baseline methods were used to compare the performance with the proposed model. The CNN model used was the standard U-Net~\citep{ronneberger2015u,cciccek20163d}, a modified tensor network (MPS TeNet) that uses one large MPS operation similar to the binary classification model in~\citep{efthymiou2019tensornetwork}, and a multi-layered perceptron (MLP). For MO-NuSeg, which is a small  dataset with 30 training examples, we also compare to nnU-Net~\citep{isensee2021nnu} which performs an extensive hyperparameter search using a five-fold cross validation and additional pre-/post- processing steps. {We followed the recommended training protocol for nnU-Net according to~\citep{isensee2021nnu} prescribed in their software repository\footnote{\url{https://github.com/MIC-DKFZ/nnunet}}}.
To compare the efficiency of the proposed method, we compare to the Lite Reduced Atrous Spatial Pyramid Pooling (LR-ASPP) network~\citep{howard2019searching} which is a MobileNet-V3 based segmentation method.  {We used the LR-ASPP pretrained on COCO 2017 dataset provided in PyTorch\footnote{\url{https://pytorch.org/vision/stable/models.html\#torchvision.models.segmentation.lraspp_mobilenet_v3_large}} without any further hyperparameter tuning, except the training learning rate.} 


Batch size (B) for different experiments were designed based on the maximum size usable by the baseline U-Net models. This resulted in the use of $1$, $32$ and $4$ for the MO-NuSeg, Lung-CXR and Retinal segmentation datasets, respectively. For the BraTS 3D dataset B was $2$. For fairer comparison all other models were trained with the same batch size. All models were trained with the Adam optimizer using an initial learning rate of $5\times 10^{-4}$, except for the MPS TeNet which required a smaller learning rate for convergence ($1\times 10^{-5}$). Training was done until 300 epochs had passed or until there was no improvement in the validation accuracy for 10 consecutive epochs. Predictions on the test set were done using the best performing model chosen based on the validation set performance. The experiments were implemented in PyTorch and trained on a single Nvidia GTX 1080 graphics processing unit (GPU) with $8$GB memory. The development and training of all models in this work was estimated to produce $71.9$ kg of CO2eq, equivalent to $601.5$ km travelled by car as measured by Carbontracker\footnote{\url{https://github.com/lfwa/carbontracker/}}~\citep{anthony2020carbontracker}.

Data augmentation was performed for the retinal vessel segmentation to alleviate the variations from the different source datasets (see Figure~\ref{fig:retina}). Horizontal flipping, rotation and affine transformations were randomly applied with $p=0.5$.

\subsubsection{Metrics} 

Segmentation performance on all datasets were compared based on Dice accuracy computed on the binary predictions, $\hat{y}_i\in\{0,1\}$, obtained by thresholding predictions at $0.5$. The threshold of $0.5$ is well suited when training with cross-entropy loss but not with Dice loss. To alleviate this we also report the area under the precision-recall curve (PRAUC or equivalently the average precision) using the soft segmentation predictions, ${s}_i\in[0,1]$ for the 2D datasets. 

The local feature map in Eq.~\eqref{eq:local} was used for MO-NuSeg and Lung-CXR datasets, whereas the feature map in Eq.~\eqref{eq:local_1} yielded better validation performance for the retinal vessel segmentation and BraTS datasets.

\subsubsection{ Model hyperparameters } The initial number of filters for the U-Net model was tuned from F=$[8,16,32]$ based on the validation performance and training time; F=$8$ for MO-NuSeg and BraTS datasets, F=$16$ for Lung-CXR , F=$32$ for retinal vessel dataset. The baseline MLP used in Lung-CXR experiments consists of 6 layers, 64 hidden units per layer and ReLU activation functions and was designed to match the strided tensor network in number of parameters. 

The strided tensor network has two critical hyperparameters: the bond dimension ($\beta$) and the stride length ($K$), which were tuned using the validation set performance. The local feature dimension ($d$) was set to $4$ for all the experiments; implications of the choice of local feature maps is presented in Section~\ref{sec:disc}.

\begin{figure}[h]
\begin{subfigure}[t]{0.49\textwidth}
    \centering
    \includegraphics[width=0.99\textwidth]{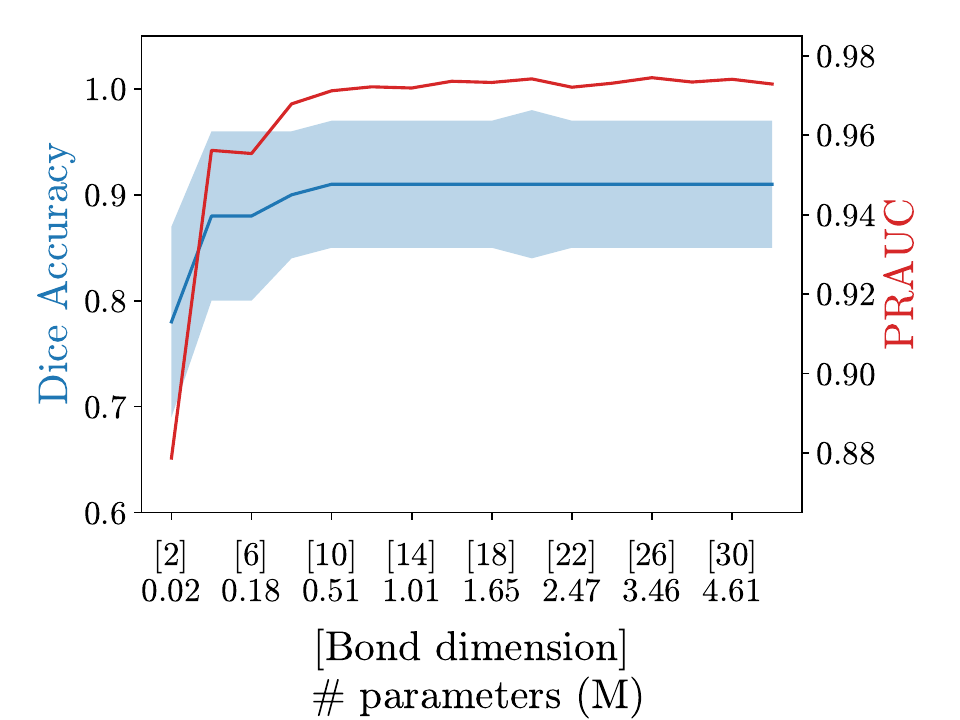}
    \caption{}
    \end{subfigure}
\begin{subfigure}[t]{0.49\textwidth}
    \centering
    \includegraphics[width=0.99\textwidth]{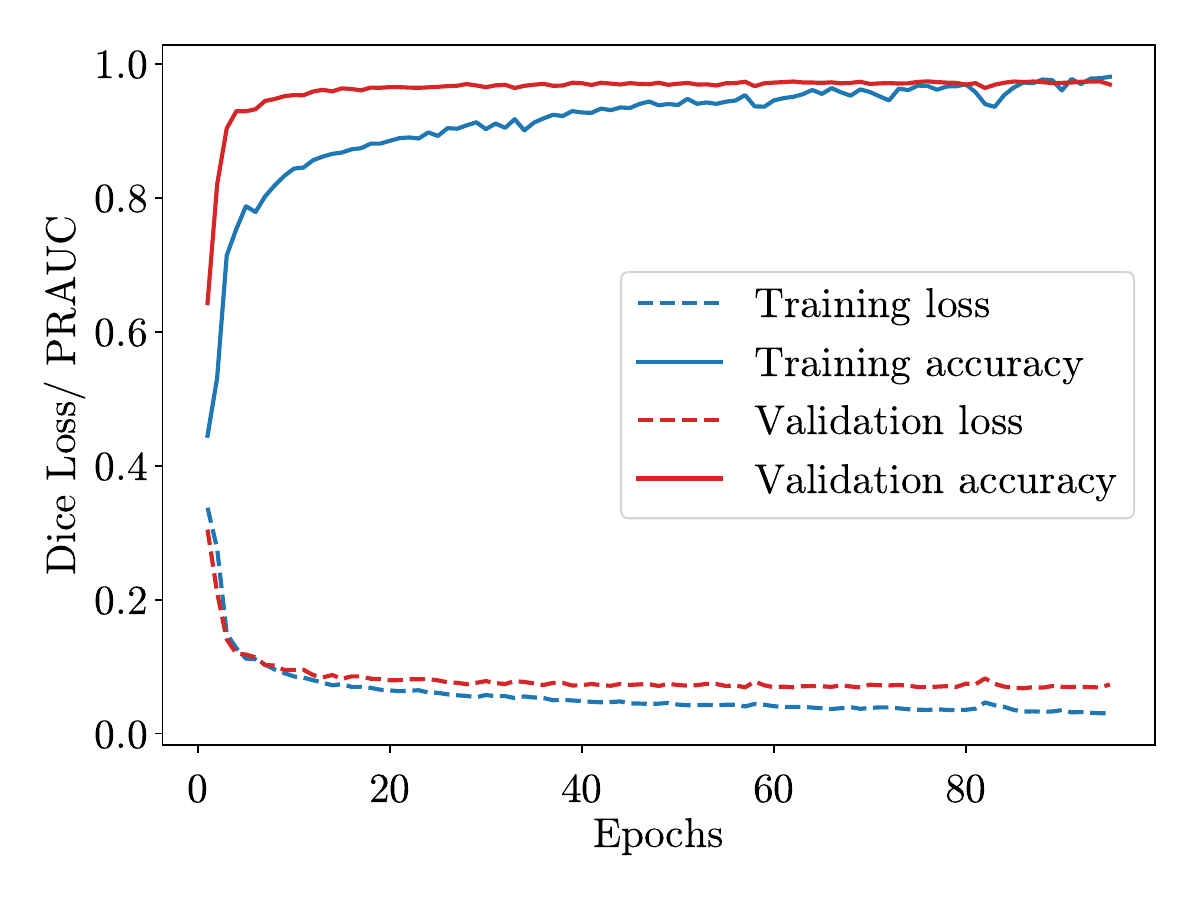}
        \caption{}
    \end{subfigure}
    \caption{a) Influence of increasing bond dimension (and hence of number of parameters) on the validation set performance for the Lung-CXR dataset. Standard deviation for Dice accuracy (blue) is computed over the validation set. b) A typical learning curve from one of the training runs for the proposed strided tenet model for the Lung-CXR dataset showing the loss and accuracy trends for the training and validation sets.}
\label{fig:learning}
\vspace{-0.25cm}
\end{figure}

\subsection{Results}

\begin{figure}[t]
    \centering
    \includegraphics[width=0.85\textwidth]{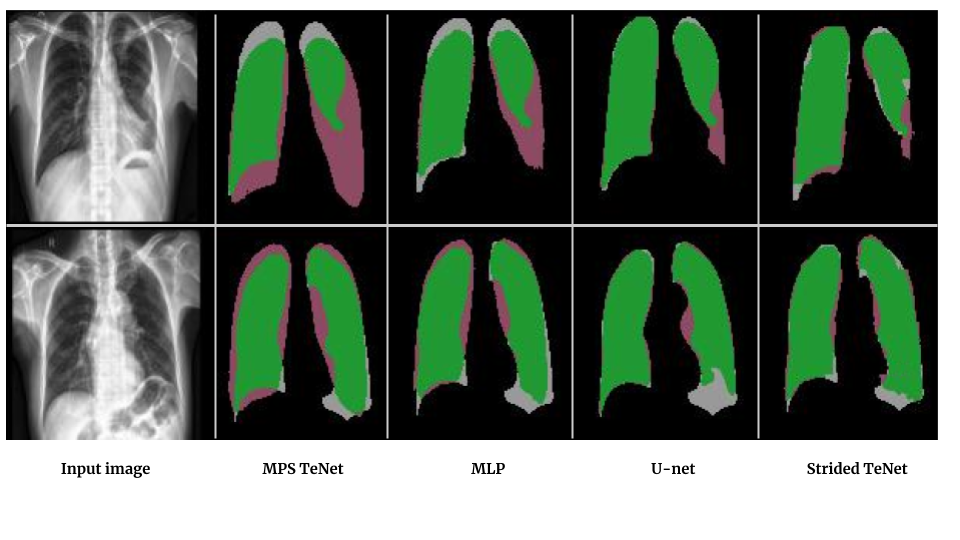}
    \caption{Two test set CXRs from Lung-CXR dataset along with the predicted segmentations from the different models. All images are upsampled for better visualisation. (Prediction Legend -- Green: True Positive, Grey: False Negative, Pink: False Positive) }
    \label{fig:cxr_res}
    \end{figure}

{\bf Tuning the bond dimension:} 
The bond dimension controls the quality of the MPS approximations and was tuned from the range $\beta=[2,4,\dots,30,32]$. The stride length controls the field of view of the MPS and was tuned from the range $K=[2,4,8,16,32,64,128]$. For MO-NuSeg dataset, the best validation performance was stable for any $\beta\geq4$, so we used the smallest with $\beta=4$ and the best performing stride parameter was $K=8$. For the Lung-CXR dataset similar performance was observed with $\beta\geq 20$, so we set $\beta=20$ and obtained $K=32$ (Figure~\ref{fig:learning}-a). The best performance on the validation set for the retinal vessel segmentation dataset was with $\beta=32$.
\begin{table}[t]
\scriptsize
\centering
    \caption{ Test set performance comparison for segmenting nuclei from the stained tissue images (MO-NuSeg) and segmenting lungs from chest CT (Lung-CXR). For all models, we report the number of parameters $|\Theta|$, computation time per training epoch, area under the curve of the precision-recall curve (PRAUC) and average Dice accuracy (with standard deviation over the test set). The representation (Repr.) used by each of the methods at input is also mentioned.}
    \label{tab:res}
     \centering
     \vspace{-0.25cm}
  \begin{tabular}{clcrrrccc}
    \toprule
    {\bf Dataset} & {\bf Models} & {\bf Repr.} &&{\bf $|\mathbf{\Theta|}$}   && {\bf t(s)} & {\bf PRAUC} & {\bf Dice} \\
    \midrule
    \multirow{4}{*}{\bf MO-NuSeg}& Strided TeNet (ours) &1D && $5.1K$&&$21.2$&$0.78$&$0.70\pm 0.10$     \\
    & U-Net &2D&&$0.5M$&& $24.5$&$0.81$&    $0.70\pm0.08$\\
    & {LR-ASPP}  &2D&& $3.2M$ && $10.5$ & $0.61$ &$0.69\pm 0.10 $\\
    & {nnU-Net} &2D&&$52.6M$&& $24.5$&$0.84$&    $0.72\pm0.12$\\
    & MPS TeNet ($\beta=10$) &1D&&$58.9M$ && $240.1$ &$0.55$&$0.52\pm0.09$\\
    & CNN  &2D&& -- && $510$ & -- &$0.69\pm 0.10 $\\
    \midrule
    \midrule
    \multirow{4}{*}{\bf Lung-CXR}&Strided TeNet (ours) & 1D&&$2.0M$&&$6.1$&$0.97$&$0.93\pm0.06$     \\
    &MLP &1D&& $2.1M$ && $4.1$&$0.95$&$0.89\pm0.05$\\
    & {LR-ASPP}  &2D&& $3.2M$ && $4.5$ & 0.63 &$0.64\pm 0.04 $\\
    &2D U-Net &2D&&$4.2M$&& $4.5$&$0.98$&    $0.95\pm0.02$\\
    &MPS TeNet ($\beta=10$)&1D && $8.2M$&&$35.7$ &$0.67$&$0.57\pm 0.09$\\
    \bottomrule
  \end{tabular}
\end{table}
\\
{\bf MO-NuSeg:}
Performance on the MO-NuSeg dataset for the strided tensor network and the baseline methods are presented in Table~\ref{tab:res} where the PRAUC, Dice accuracy, number of parameters and the average training time per epoch are reported. The proposed Strided TeNet (PRAUC=$0.78$, Dice=$0.70$) model obtains comparable performance with U-Net (PRAUC=$0.81$, Dice=$0.70$). There was no significant difference between the two methods in their Dice accuracy based on paired sample t-tests. {We also report the performance of nnU-Net  which shows a small gain in performance (PRAUC=$0.84$, Dice=$0.72$) compared to the standard U-Net we report}.
We report the performance of the MPS TeNet model in~\citet{efthymiou2019tensornetwork} which operates on flattened input images obtaining a Dice score of $0.52$ and PRAUC of $0.55$. {We further compare to the efficient LR-ASPP network which obtains a Dice score of $0.64\pm0.10$ and PRAUC of $0.61$}.
Finally, we reproduce the CNN based method from the MO-NuSeg dataset paper in ~\citet{kumar2017dataset} where they reported a Dice accuracy of $0.69\pm 0.10$.\footnote{The run time with more recent hardware can be lower than what is reported in~\citet{kumar2017dataset}. Note that the exact training-test splits between our experiments and~\citet{kumar2017dataset} might not be identical.}
\\
{\bf Lung-CXR dataset:} Performance measures on the Lung-CXR dataset are also reported in Table~\ref{tab:res}. As this is a relatively simpler dataset both U-Net and Strided TeNet models obtain very high PRAUC: $0.98$ and $0.97$, respectively. There is a difference in Dice accuracy between the two: $0.95$ for U-Net and $0.93$ for Strided TeNet, without a significant difference based on paired sample t-test. In addition to the MPS TeNet (Dice=$0.57$) we also report the performance for an MLP which obtains Dice score of $0.89$. {We also compare to the efficient LR-ASPP network which obtains a Dice score of $0.64\pm0.04$ and PRAUC of $0.63$.}
Qualitative results based on two predictions where the strided tensor network had high false positive (top row) and high false negative (bottom row), along with the predictions from other methods and the input CXR are shown in Figure~\ref{fig:cxr_res}. 
\begin{table}[h]
\scriptsize
\centering
    \caption{ Performance comparison for segmenting retinal vessels in the Retinal dataset. For all models, we report the number of parameters $|\Theta|$, computation time per training epoch, area under the curve of the precision-recall curve (PRAUC) and average Dice accuracy (with standard deviation over the test set).}
    \label{tab:ret_res}
     \centering
  \begin{tabular}{lcrrrccc}
    \toprule
    {\bf Models} &&{\bf $|\mathbf{\Theta|}$}   && {\bf t(s)} & {\bf Dice} \\
    \midrule
    2D U-Net &&$7.5M$&& $5.2$&    $0.67\pm0.16$\\
    Strided TeNet (ours) && $3.2M$&&$4.5$&$0.64\pm 0.10$     \\
    Hybrid Strided TeNet && $2.3M$&&$3.8$&$0.66\pm 0.10$     \\
    MPS TeNet ($\beta=10$) &&$11.4M$ && $26.1$ &$0.31\pm0.04$\\
    {LR-ASPP} &&$3.2M$ && $8.1$ &$0.23\pm0.15$\\
    \bottomrule
  \end{tabular}
\end{table}
\begin{figure}[h]
    \centering
    \includegraphics[width=0.95\textwidth]{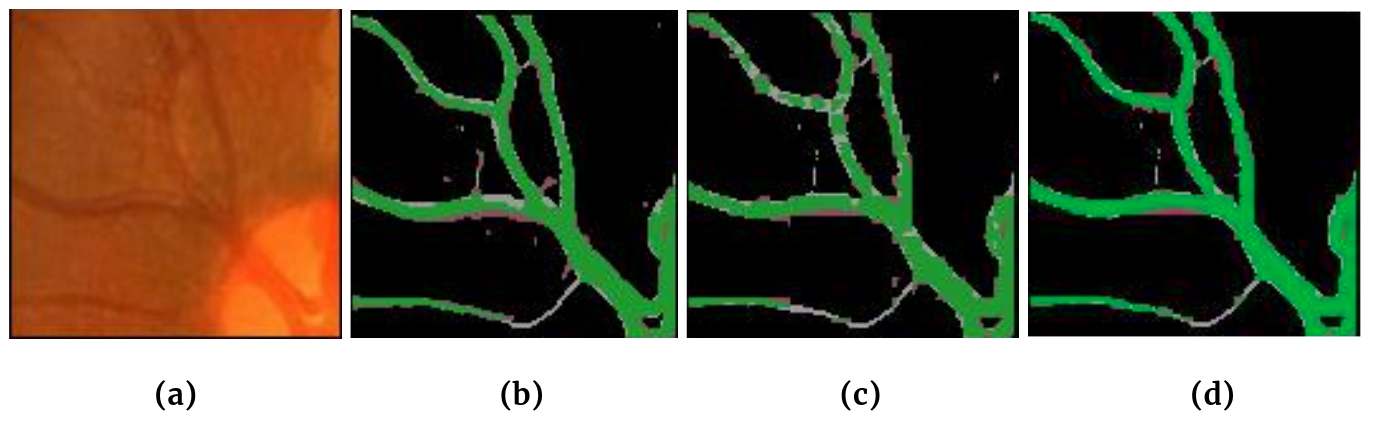}
    \caption{Visualisation of the predictions from U-Net (b) and Strided TeNet (c) on one retinal image patch (a). The false negative errors between the two methods are of different nature; Strided TeNet misses block like regions (seen clearly when zoomed in) due to the striding over $K\times K$ regions. (d) Predicted segmentation from the Hybrid Strided TeNet which uses a 2-layered CNN-based local feature map. (Prediction Legend -- Green: True Positive, Grey: False Negative, Pink: False Positive) }
    \label{fig:ret_res}
    \end{figure}
\\
{\bf Retinal vessel segmentation dataset:} Results for the test set for segmenting retinal vessels is reported in Table~\ref{tab:ret_res} where we notice that U-Net (Dice=$0.67$) shows an improvement compared to the Strided TeNet (Dice=$0.64$). However, using a 2-layered CNN based local feature map and the MPS reported as the Hybrid Strided TeNet in Section~\ref{sec:local}  is able to bridge the gap in performance compared to U-Net. The MPS TeNet is unable to handle the complexity of this dataset (Dice=$0.31$). {The efficient LR-ASPP network struggles on this dataset and obtains a Dice score of $0.23\pm0.15$}. Qualitative results are visualized in Figure~\ref{fig:ret_res} for one input patch for U-Net and Strided TeNet.
\\
{\bf BraTS dataset:} The Strided TeNet was primarily formulated as a 2D segmentation model. With these experiments on the BraTS dataset we study its performance on 3D segmentation task by using flattened 3D patches instead of 2D patches as input to the Strided TeNet. The results are reported in Table~\ref{tab:brats_res}. We compare its performance with 3D U-Net~\citep{cciccek20163d} and see that U-Net (Dice=$0.69$) easily outperforms the Strided TeNet (Dice=$0.62$). 

\begin{table}[t]
\scriptsize
\centering
    \caption{ Test set performance for the 3D segmentation task on BraTS dataset. {We report the number of parameters $|\Theta|$, computation time per training epoch and average Dice accuracy (with standard deviation over the test set).}}
    \label{tab:brats_res}
     \centering
  \begin{tabular}{lcrrrccc}
    \toprule
     {\bf Models} &&{\bf $|\mathbf{\Theta|}$}   && {\bf t(s)} & {\bf Dice} \\
    \midrule
    Strided TeNet (ours)  && $1.3M$&&$70$&$0.63\pm 0.05$     \\
    3D U-Net&&$1.4M$&& $320$&    $0.69\pm0.08$\\
    \bottomrule
  \end{tabular}
\end{table}

\section{Discussion}
\label{sec:disc}
\subsection{Performance on 2D data}

Results from Table~\ref{tab:res} for MO-NuSeg and Lung-CXR datasets show that the proposed strided tensor network compares favourably to other baseline models. In both cases, there was no significant difference in Dice accuracy compared to U-Net. The computation time per epoch for the strided tensor network was also in the same range as that for U-Net. The total training time for U-Net was smaller as it converged faster ($\approx 50$ epochs) than strided tensor network model ($\approx100$ epochs). However, in all instances the strided tensor network converged within one hour. Also, note that all the models in this work were implemented in PyTorch~\cite{paszke2019pytorch} which do not have native support for tensor contraction operations. The computation cost for tensor networks can be further reduced with the use of more specialized implementations such as the ones being investigated in recent works like in~\citep{fishman2020itensor,novikov2020tensor}. 
{Additionally, performance with nnU-Net~\citep{isensee2021nnu} on the MO-NuSeg dataset in Table~\ref{tab:res} shows that nnU-Net achieves a higher Dice score. However, this small improvement could be attributed due to the additional pre- and post- processing conducted in the nnU-Net pipeline, which is not followed for any other reported methods.}

The qualitative results in Figure~\ref{fig:cxr_res} reveal interesting behaviours of the methods being studied. The predicted segmentations from the MPS TeNet (column 2) and MLP (column 3) are highly regularised that resemble {\em average} lung representations. This can be attributed to the fact that both these models operate on 1D inputs (flattened 2D images); this loss of structure forces the model optimisation to converge to a mode that seems to capture the average lung mask. In contrast, the predictions from U-Net (column 4) and strided tensor network (column 5) are closer to the ground truth and are able to capture variations between the two input images. This can be explained by that fact that U-Net directly operates on 2D input and the underlying convolutional filters learn features that are important for the downstream lung segmentation. The strided tensor network, on the other hand, operates on 1D inputs but from smaller image patches. The strided tensor network is able to overcome the loss of spatial structure compared to the other models operating on 1D inputs by using weight-shared MPS on small patches.

The results for retinal vessel segmentation task are reported in Table~\ref{tab:ret_res}. Compared to the first two 2D datasets that comprise structures of interest that are largely regular in shape, this dataset is more challenging due to the variations in scale, shape and orientation of the vessels. This is further aggravated by the differences in acquisition between the three data sources as described in Section~\ref{sec:data}. The Dice accuracy for U-Net ($0.67$) shows a small improvement that is significant compared to the strided tensor network ($0.64$) on this dataset. It can be explained by the large U-Net model that was used for this task ($7.5$M parameters) compared to strided tensor network ($3.2$M). Further, and more importantly, the 1D input to the strided tensor network could affect its ability to segment vessels of varying sizes, as seen in Figure~\ref{fig:retina}. 

\subsection{Influence of local feature maps}

The sinusoidal local feature maps in Equations~\eqref{eq:local} and~\eqref{eq:local_1} are used to increase the local dimension ($d$), so that the lift to high dimensions can be performed. Non-linear pixel transformations of these forms are also commonly used in many kernel based methods and have been discussed in earlier works~\citep{stoudenmire2016supervised}. More recently, scalable Fourier features of the form in Eq.~\eqref{eq:local_1} have been extensively used in attention based models operating on sequence-type data to encode positional information~\citep{vaswani2017attention,mildenhall2020nerf,jaegle2021perceiver}. Although we do not explicitly encode positional information, the use of these scalable Fourier features in Eq.~\eqref{eq:local_1} yielded improved results in the retinal vessel segmentation task compared to use the local feature maps in Eq.~\eqref{eq:local}. 

CNNs are widely used as feature extractors for their ability to learn features from data that are useful for downstream tasks. Building on this observation, we use CNNs to {\em learn} local features instead of the non-informative sinusoidal features. We attempt this by using a simple 2-layered CNN with 8 filters in each layer; the output features from the CNN form the local feature map which are then input to the strided tensor network. The use of CNN along with the strided tensor network results in what we call the {\em hybrid} strided tensor network. Performance for this hybrid model (Dice=$0.66$) is reported in Table~\ref{tab:ret_res} where we see it gives a small improvement over the standard strided tensor network (Dice=$0.64$). The scope of this work was to explore the effect of learnable feature maps and it demonstrated to be beneficial. Further investigations into how best to combine the strengths of CNNs as feature extractors and tensor networks for approximating decision boundaries could yield a powerful class of hybrid models.

\subsection{Resource utilisation}
Recent studies point out that overparameterisation of deep learning models are not necessarily detrimental to their generalization capability~\citep{allen2019learning}. However, training large, overparameterised models can be resource intensive~\citep{anthony2020carbontracker}. To account for the difference in resource utilisation we compare the number of parameters for each of the models in Table~\ref{tab:res}. From this perspective, we notice that for the MO-NuSeg data, the number of parameters used by the strided tensor network ($5.1K$) is about two orders of magnitude smaller than that of the U-Net ($500K$) without a substantial difference in segmentation accuracy. As a consequence, the maximum GPU memory utilised by the proposed tensor network was $0.8$GB and it was $6.5$GB for U-Net.  { The number of parameters for nnU-Net ($52.6M$) was four orders of magnitude larger than that of the Strided TeNet. The additional resource consumption due to the extensive cross-validation based model selection in nnU-Net further aggravates the increase in compute resources (for instance, nnU-Net required about 60 hours to perform model selection on MO-NuSeg with only 30 training examples)}.

Another reason for reduced GPU memory utlisation for the strided tensor network is that they do not have intermediate feature maps and do not store the corresponding computation graphs in memory, similar to MLPs~\citep{raghav2020tensor}. This difference in resource utlisation can be advantageous for strided tensor network as training can be carried out with larger batches which can further lower the computation time and more stable model training~\citep{keskar2017}. To demonstrate these gains in resource utlisation, we trained the strided tensor network with the largest batch size possible:  for the MO-NuSeg where images were of size $1000\times1000$, it was $B=12$ (whereas $B=1$ for U-Net) without any degradation in performance. {Methods such as the LR-ASPP from~\citet{howard2019searching} focus on improving the efficiency of CNN-based segmentation methods by improving upon existing MobileNet architectures. {MobileNet architectures introduced depth-wise separable convolutions by separating spatial filtering from feature generation to reduce the computation complexity of performing convolutions. This has some similarity with the MPS operations where the pixel classification is performed on small patches without feature extraction. However, one main difference with our Strided TeNet is that we do it in a single layer unlike LR-ASPP or other MobileNet architectures which use {\em deep}  architectures.}  Our experiments showed these methods to be efficient in terms of GPU memory utilization and training times. However, the performance of LR-ASPP across all 2D datasets was lower than the Strided TeNet. Fine grained tuning of model hyperparameters and pre-training strategies could improve the performance of LR-ASPP.}
 
 \begin{figure}[t]
    \centering
    \vspace{-0.25cm}
    \includegraphics[width=0.65\textwidth]{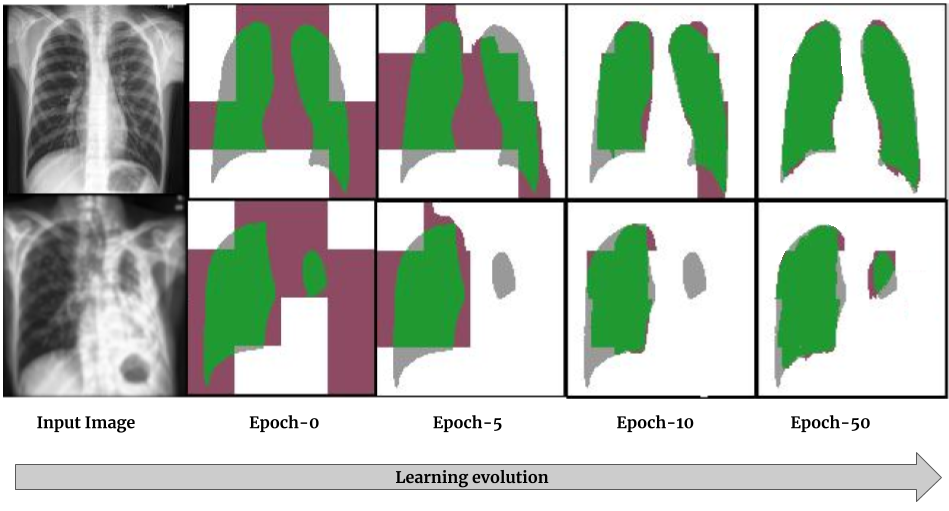}
    \caption{Progression of learning for the strided tensor network, for the task of segmenting lung regions. Two {\em validation} set input chest CT images (column 1) and the corresponding predictions at different training epochs are visualized overlaid with the ground truth segmentation. Images are of size 128$\times$128 and the stride is over 32$\times$32 regions. All images are upsampled for better visualisation. (Prediction Legend -- Green: True Positive, Grey: False Negative, Pink: False Positive) }
    \label{fig:evolution}
    \vspace{-0.5cm}
    \end{figure}

\subsection{MPS as a learnable filter} 
CNN-based models such as the U-Net operate by learning filter banks that are able to discern features and textures necessary for a given downstream task. In contrast, the proposed strided tensor network learns to classify pixels into foreground and background classes by operating on small flattened image patches using a single weight-shared MPS tensor network. In CNN jargon, this is analogous to learning a single filter that can be applied to patches from across the image. This behaviour of the strided tensor network for segmenting lungs by operating on patches of size $32\times32$ and the gradual evolution of the MPS {\em filter} is illustrated in Figure~\ref{fig:evolution}. Initially, the predictions are at the patch level where either an entire $K\times K$ patch is either predicted to be foreground or background. Within a few epochs the tensor network is able to distinguish image features to predict increasingly more accurate lung masks.

\subsection{{Influence of stride length (K)}}

%
%
{
The proposed Strided TeNet is shallow (in fact, single layered) and performs segmentation at the input resolution of the images. This is in contrast with deep (multi-scale) CNN-based methods such as the U-Net~\citep{ronneberger2015u}. 
As discussed in Sec.~\ref{sec:data_exp}, stride length is an important hyperparameter of our method and for the various experiments it was tuned from the range, $K=[2,4,8,16,32,64,128]$. The optimal $K$ varied depending on the structures of interest in each of the datasets. For instance, when segmenting small nuclei in MoNuSeg dataset $K=8$ whereas for the lung segmentation task $K=32$. In effect, the stride length directly controls the receptive field of our method. In cases where the structures of interest could be of varying sizes -- for instance in multi-organ segmentation tasks -- a single stride length might not be optimal. One way to overcome this could be by using a one-vs-all segmentation model where each class is segmented by a binary strided tensor network with a specific stride length. Due to the overall lower resource requirements for strided tensor networks, this combined approach could still be a competitive method.}

\subsection{Performance on 3D data}
At the outset, the strided tensor network model was primarily designed for 2D segmentation tasks. However, the fundamental principle of how the segmentation is performed -- by lifting small image regions to high-dimensional spaces and approximating linear classification -- could be directly translated to 3D datasets. With this motivation we performed the experiments on the BraTS dataset for segmenting tumour regions from 3D MRI scans, where the high-dimensional lift was performed on flattened 3D patches of size $K^3$ instead of 2D patches. The results in Table~\ref{tab:brats_res} indicate that the strided tensor network approach does not immediately translate to 3D datasets. 

There are two possible reasons for these negative results. Firstly, the loss of structure going from 3D to 1D hampers segmentation more than in 2D data. In previous work in~\citet{selvan2020locally} where tensor networks were used to perform image classification on 3D data, a multi-scale approach was taken. A similar approach that integrates multiple scales as in an encoder-decoder type model such as the 3D U-Net perhaps could be beneficial. Secondly, the number of parameters required to learn the segmentation rules in 3D data could be far higher than the 2D models which were applied on the 3D dataset. Due to GPU memory constraints we could not explore larger tensor network models with higher bond dimensions ($>12$) for 3D data, as we found it to be beneficial for the more challenging vessel segmentation task where bond dimension was set to $32$. Potentially, these challenges could be overcome by taking up a multi-planar approach similar to the multi-planar U-Net~\cite{perslev2019one} where the training is performed using multiple 2D planes and the final 3D segmentation is fused from these views.

\section{Conclusions}
 
We presented a novel tensor network based medical image segmentation method which is an extension of our preliminary work in~\citet{selvan2021segmenting}. We proposed a formulation that uses the matrix product state tensor network to learn decision hyper-planes in exponentially high-dimensional spaces. The high-dimensional lift of image pixels is performed using local feature maps, and we presented a detailed study of the implications of different parametric and non-parametric local feature maps. The loss of spatial correlation due to the flattening of input images was overcome by using weight-shared MPS that operate on small image patches. This we also demonstrated to be useful in reducing the exponential increase in the number of parameters. The experiments on three 2D segmentation datasets demonstrated promising performance compared to popular deep learning methods. Further, we adapted the proposed strided tensor network for segmenting 3D data. Although these experiments resulted in negative results, the insights gained by these experiments were discussed and ideas for further improvements have been presented. The general idea of using simple local feature maps, lifting the data to exponentially high-dimensional feature spaces and then approximating segmentation decisions using linear models like the tensor networks is a novel class of methods that can be explored further. 


\bibstyle{plain}
\setlength{\bibsep}{5pt}
\small
\bibliography{selvan20}


\end{document}